\documentclass[letterpaper, 10 pt, conference]{ieeeconf}  

\IEEEoverridecommandlockouts                              

\usepackage{graphics} 
\usepackage{epsfig} 
\usepackage{amsmath} 
\usepackage{amssymb}  
\usepackage{siunitx}
\usepackage[percent]{overpic}
\makeatletter\newcommand{\manuallabel}[2]{\def\@currentlabel{#2}\label{#1}}\makeatother
\usepackage{hyperref}

\usepackage{fancyhdr}
\fancypagestyle{plain}{%
	\fancyhf{} 
	\fancyfoot[C]{ © 2020 IEEE.  Personal use of this material is permitted.  Permission from IEEE must be obtained for all other uses, in any current or future media, including reprinting/republishing this material for advertising or promotional purposes, creating new collective works, for resale or redistribution to servers or lists, or reuse of any copyrighted component of this work in other works.}
	
}
\title{\LARGE \bf
Variable Autonomy of Whole-body Control for Inspection and Intervention in Industrial Environments using Legged Robots
}

\author{Guiyang Xin, Carlo Tiseo, Wouter Wolfslag, Joshua Smith, Oguzhan Cebe,\\ Zhibin Li, Sethu Vijayakumar, Michael Mistry
\thanks{*This work is funded by the EPSRCs RAI Hubs for Extreme and Challenging Environments: NCNR EPR02572X/1 and ORCA EPR026173/1, as well as the European Commission Horizon 2020 WorkProgramme: THING ICT-2017-1 780883 and MEMMO ID: 780684.}
\thanks{All the authors are with the Institute for Perception, Action and Behaviour, School of Informatics, The University of Edinburgh.
        {\tt\small guiyang.xin@ed.ac.uk}}%
}

\begin{document}

\maketitle
\thispagestyle{fancy}%


\begin{abstract}
The deployment of robots in industrial and civil scenarios is a viable solution to protect operators from danger and hazards. Shared autonomy is paramount to enable remote control of complex systems such as legged robots, allowing the operator to focus on the essential tasks instead of overly detailed execution.
To realize this, we propose a comprehensive control framework for inspection and intervention using a legged robot and validate the integration of multiple loco-manipulation algorithms optimised for improving the remote operation. 
The proposed control offers 3 operation modes: fully automated, semi-autonomous, and the haptic interface receiving onsite physical interaction for assisting teleoperation.
Our contribution is the design of a QP-based semi-analytical whole-body control, which is the key to the various task completion subject to internal and external constraints. We demonstrate the versatility of the whole-body control in terms of decoupling tasks, singularity tolerance and constraint satisfaction. We deploy our solution in field trials and evaluate in an emergency setting by an E-stop while the robot is clearing road barriers and traversing difficult terrains.
\end{abstract}

\section{INTRODUCTION}
The increase of the human life value over the last two centuries has been unprecedented, considering that few decades ago what is now considered child labor was a reality in all the G7 countries. The increase in population wealth is not only related with the perceived value of human life but it is also related to the increase life expectancy and the minimization of infant mortality. Therefore, the risk associated with jobs such as working on offshore platforms and nuclear sites is becoming unaffordable in the modern society. 

Projects as ORCA (Offshore Robotics for Certification of Assets, \url{https://orcahub.org/}) aim to develop, test and validate robotic technology that in the foreseeable future can be deployed along side the human operators to minimise the work related perils. Legged robots can help in such sites because they offer more maneuverability than wheeled robots in real world scenarios, making them easier to deploy in traditional human operated industrial facilities. However, this comes at the cost of limited payloads and lower reliability due to possible and more difficult locomotion stability, which lead ORCA to research also drones and wheeled platforms. 

\begin{figure}[t!]
\centering
 \includegraphics[width=\linewidth]{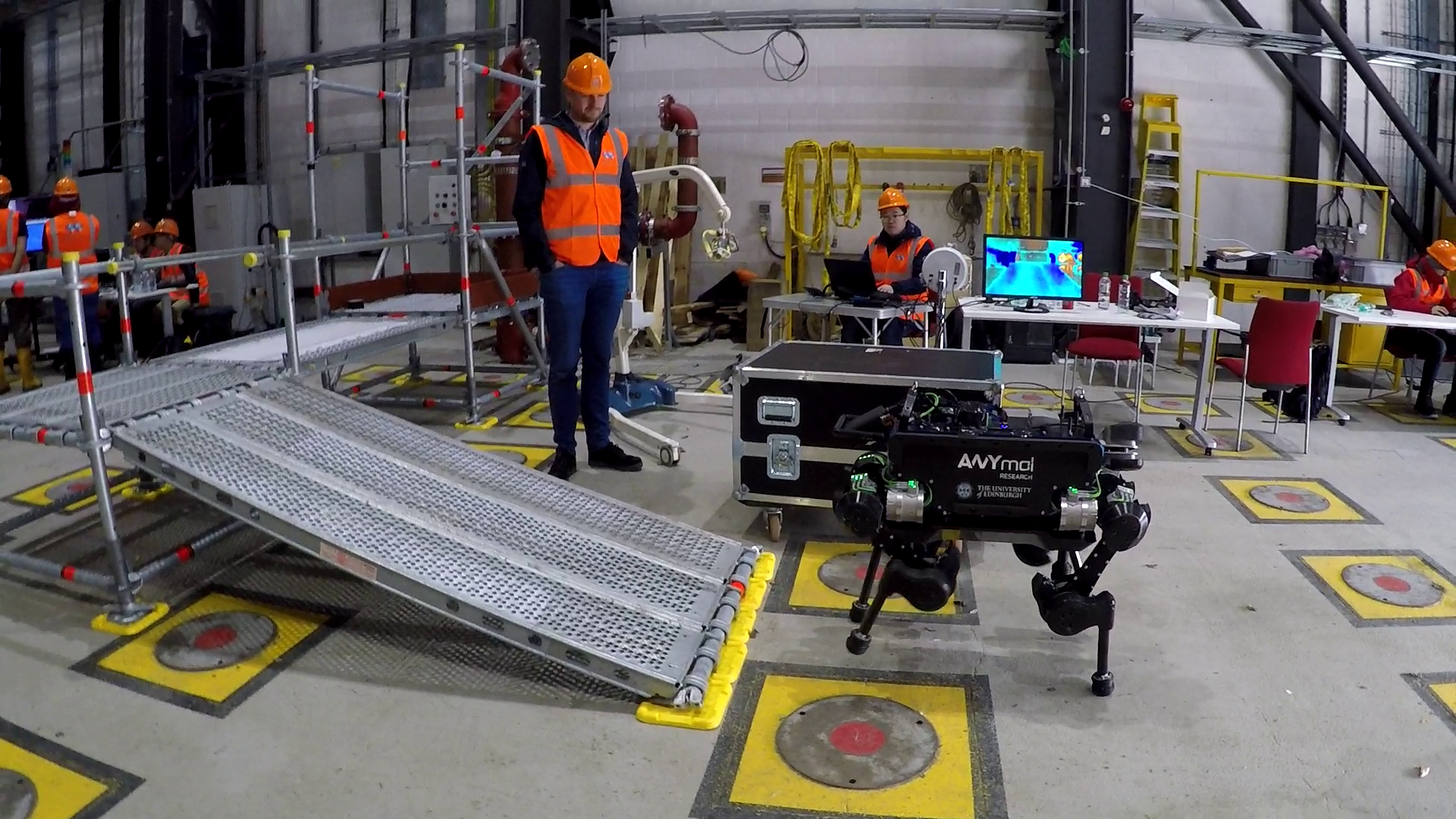}
 \caption{ORCA project testing: a quadruped robot was teleoperated to traverse industrial scaffolding, and eventually pressed an E-stop button, while thermal information was fed-back to the operator.}
 \label{f:fig1}
\end{figure}

An integral part of our research within the ORCA project is testing our methods in industrial environments, where we conduct regular demonstrations  attended by leading companies in the offshore industry. Feedback from previous interactions with these companies helped us develop a comprehensive obstacle course to showcase our controllers for a comprehensive routine inspection with some intervention capabilities as shown in Fig. \ref{f:fig1}. The course requires the removal of an obstacle to access to the scaffolding, which is composed of a slope up, a cluttered \SI{90}{\degree} turn, a slippery bridge leading to a control box where the remote operator uses one of the legs to push an E-Stop button. 

To achieve such a routine we integrate different planners with a whole-body controller developed for floating base systems. Planners generate the foot step sequence and trajectories of feet and base for locomotion and manipulation, feeding the instantaneous whole-body controller via interpolations. The demonstration of this paper shows the effectiveness of our system and techniques regarding to loco-manipulation in real industrial applications. 

\subsection{Related work}

Autonomous mobile manipulation has been a research hot-spot since DARPA Robotics Challenge (DRC), with subsequent international competitions such as DARPA Subterranean Challenge and Mohamed Bin Zayed International Robotics Challenge (MBZIRC). Legged robots are a particularly challenging agent to accomplish these competitive tasks due to the control complexity of a multiple degrees of freedom (DOFs) robot and the inherent risk of falling. Loco-manipulation control allows the robot to coordinate multiple tasks simultaneously and also to tackle any interference of tasks due to potential lack of DOFs.

Early work to coordinate multiple tasks are introduced in the kinematic level \cite{mistry2008inverse}, by solving inverse kinematics and execution by single joint controllers. Obviously, kinematic coordination cannot account for dynamic decoupling and satisfy dynamic constraints. Optimization-based whole-body controllers are proposed to overcome the drawback of single joint control. \cite{Righetti2013} formulates internal and external constraints into a quadratic programming (QP) problem with optimized contact forces. A widely used contact force optimization method called the Virtual Model Controller \cite{gehring2013control}\cite{boaventura2012dynamic} has to control the swing legs using joint PD controllers. Therefore, it is not a pure torque-based whole-body controller. \cite{de2009prioritized}\cite{de2010feature} firstly proposed the hierarchical quadratic programming (HQP) approach to solve each task in prioritized QPs with full dynamic constraints sequentially. \cite{kanoun2011kinematic} also formulates a cascade of QP to involve in inequality constraints for humanoid robot control. Herzog et al. \cite{Herzog2016} combine the two benefits of having inequalities in all hierarchical levels \cite{kanoun2011kinematic} and reducing the number of variables from one QP to the other \cite{de2010feature}. \cite{del2015prioritized} compares the computation efficiency of different HQP formulations. Beyond those instantaneous whole-body controller, optimal control techniques \cite{budhiraja2018differential}\cite{mastalli2019crocoddyl} are proposed to generate optimal control command along with optimal trajectories. But optimal control is too slow for real time control when considering inequality constraints. A more general framework employing nonlinear programming (NLP) is proposed to cope with planning and control for loco-manipulation of low dimension mobile robots \cite{merkt2019towards} or offline motion planning \cite{ferrolho2020optimizing}.

Compared to optimization-based controllers, inverse dynamic controllers can give us more analytical insights of systems. Khatib \cite{khatib1987unified} derives the operational space control (OSC), establishing the straightforward relationship between operational tasks and inverse dynamics. OSC has been extended to hierarchical OSC by adding iterative null-space projections for legged robots \cite{sentis2010compliant}\cite{lee2016balancing}\cite{sentis2006whole}. Aghili \cite{aghili2005unified} proposes the projected inverse dynamic control scheme to decouple constraint and motion in a constrained system. Mistry et al. \cite{mistry2012operational} applies the projection matrix to underacted systems and implements OSC in constraint-free space. Satisfying inequality constraints, particularly friction cone, is crucial to legged robots because of the interaction with environments. In \cite{Xin2018}, we extend that approach to combine analytical Cartesian impedance controller proposed in \cite{albu2003cartesian} and QP optimization aimed at full-filling inequality constraints. This QP-based semi-analytical controller benefits on computation efficiency and gives us the ability to estimate contact forces leveraging Cartesian impedance control \cite{xin2020optimization}.

\subsection{Contributions}

In this work, we apply our control framework to implement a sequence of locomotion and manipulation tasks in industrial environments. We evaluate our proposed system by demonstrating a whole-body motion (moving base with fixed end-effector posture) showcasing the decoupling of operational space foot posture control from base motion. We also demonstrate the controller can handle body-ground contact constraints without any adaptation. We highlight the great potential of a quadruped robot running our controller in real world applications. 
Figure \ref{f:fig1} shows the testing site, and accompanying video is available at: \url{https://youtu.be/tIyfUjJgJIM}.

\section{System overview}

Our system consists of a quadruped robot ANYmal\footnote{https://www.anybotics.com} with a RGB camera and a thermal camera for perception. A HRI\footnote{http://humanisticrobotics.com} remote controller is used to send walking velocities of the base frame attached on the torso, meanwhile a Sigma 7\footnote{https://www.forcedimension.com} haptic joystick is for teleoperating a foot to do manipulation. The GUI integrates camera windows and interfaces for controller switching and parameter handling. One on-board PC is running planners \cite{fankhauser2016free} and controllers plus a state estimator within 400 Hz cycle. Another on-board PC is connected to two cameras, sending perception images to the GUI. Each on-board computer has an Intel 4th generation (Haswell ULT) i7-4600U (1.4 GHz-2.1 GHz) processor and two HX316LS9IBK2/16 DDR3L memory cards. The operation commands are wrapped as ROS messages to communicate with on-board systems via wireless network. The diagram of Fig. 2 shows the modules of the system.

\begin{figure}[t!]
\centering
 \includegraphics[width=\linewidth]{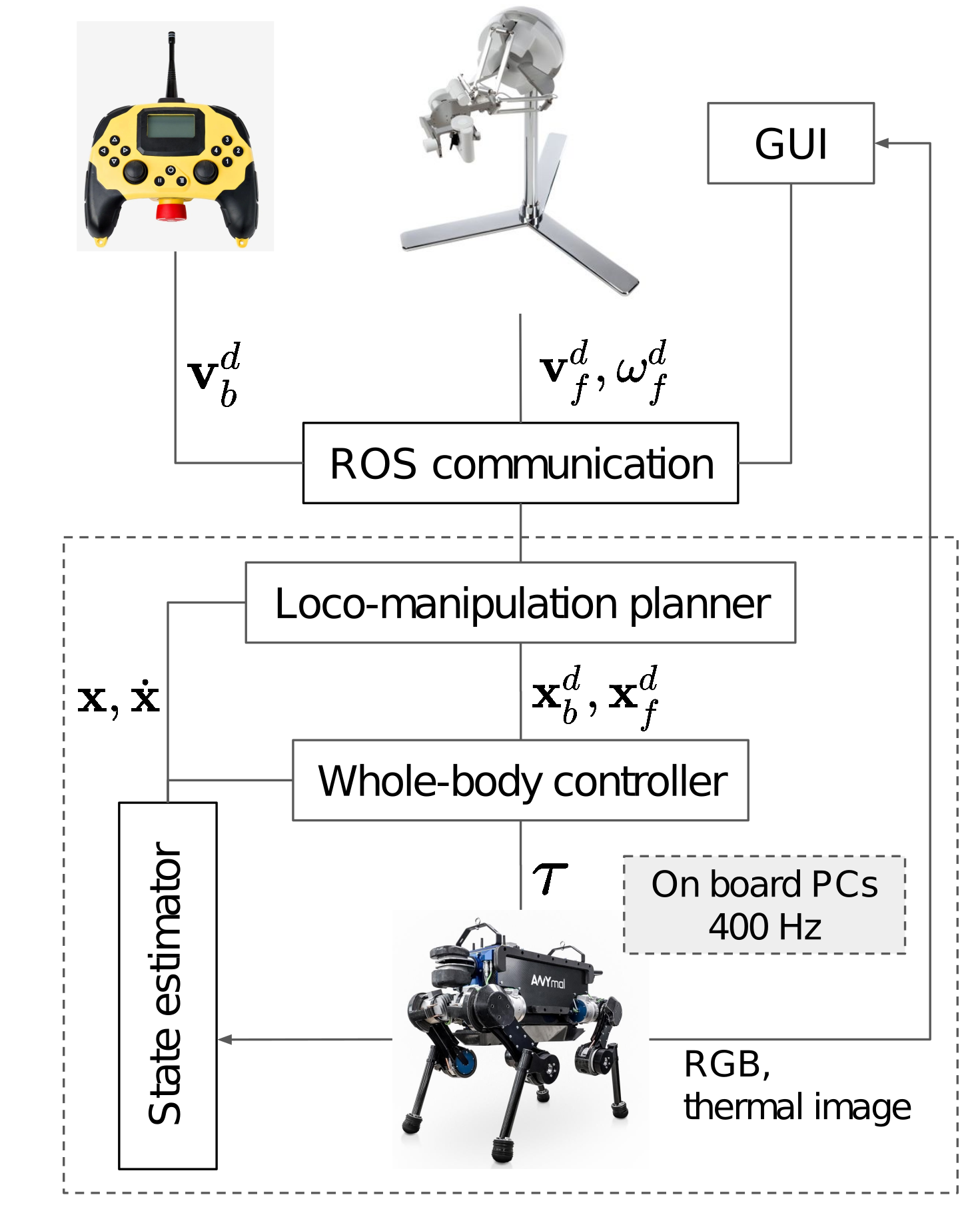}
 \caption{System overview. Loco-manipulation planner interprets signals from two joysticks and generates task space trajectories and foot steps. Whole-body controller executes the desired trajectories and also satisfies environment constraints. Operators manipulate the joysticks with assistance of camera visions.}
 \label{f:system}
\end{figure}

\section{Methods}

The goal of whole-body control is to generate torque commands for all the actuators corresponding to desired tasks and subject to physical constraints. Since there are inequality constraints such as friction cones, the whole-body controller has to be an optimization-based controller. In our case, we use a semi-analytical QP-based controller described in \cite{Xin2018} with the benefits of efficient computation and contact force estimation \cite{xin2020bounded} compared to other purely optimization-based controller such as HQP controllers \cite{Herzog2016}\cite{DarioBellicoso2016}. We use the same controller for locomotion, operation and body-ground contact scenario. Particularly, body-ground contact scenario was never explored in terms of whole-body control except our paper \cite{wolfslag2020iros}. Here we show that we can use the same control strategy to handle constraints acting on feet and body. 

\subsection{Model formulation}

The configuration of a floating base system with limbs is fully determined by a generalized coordinate vector $\mathbf{q}=\begin{bmatrix}{}_{I}\mathbf{x}_b^\top &\mathbf{q}_j^\top\end{bmatrix}^\top \in SE(3) \times n$, where ${}_{I}\mathbf{x}_b \in SE(3)$ denotes the floating base's position and orientation with respect to a fixed inertia frame, meanwhile $\mathbf{q}_j \in \mathbb{R}^n$ denotes the vector of actuated joint positions. Also, we define the generalized velocity vector as $\dot{\mathbf{q}}=\begin{bmatrix}{}_{I}\mathbf{v}_b^\top & {}_{B}\boldsymbol{\omega}_b^\top &\dot{\mathbf{q}}_j^\top\end{bmatrix}^\top \in \mathbb{R}^{6+n}$, where ${}_{I}\mathbf{v}_b \in \mathbb{R}^3$ and ${}_{B}\boldsymbol{\omega}_b \in \mathbb{R}^3$ are the linear and angular velocities of the Base with respect to the inertia frame expressed respectively in the $I$ and $B$ frame. The equations of motion of a floating base system subject to environment contacts are written as 
\begin{equation}\label{e:dynamics}
      \mathbf{M}(\mathbf{q})\mathbf{\ddot{q}}+\mathbf{h}(\dot{\mathbf{q}},\mathbf{q})=\mathbf{B}\boldsymbol{\tau}+\mathbf{J}_c^{\top}(\mathbf{q})\boldsymbol{\lambda}_c
\end{equation}
where $\mathbf{M}(\mathbf{q}) \in \mathbb{R}^{(n+6)\times(n+6)}$ is the inertia matrix, $\mathbf{h}(\dot{\mathbf{q}},\mathbf{q}) \in \mathbb{R}^{n+6}$ is the generalized vector containing Coriolis, centrifugal and gravitational effects,
$\boldsymbol{\tau} \in \mathbb{R}^{n+6}$ is the vector of torques, $\mathbf{J}_c(\mathbf{q}) \in \mathbb{R}^{3k\times(n+6)}$ is the constraint Jacobian that describes $3k$ constraints, $k$ denotes the number of contact points accounting foot contact and body contact,
$\boldsymbol{\lambda}_c \in \mathbb{R}^{3k}$ are constraint forces acting on contact points, and
\begin{equation}\label{B}
    \mathbf{B}=\begin{bmatrix}
        \mathbf{0}_{6 \times 6} & \mathbf{0}_{6 \times n}   \\
        \mathbf{0}_{n \times 6}   & \mathbf{I}_{n \times n}
    \end{bmatrix}
\end{equation}
is the selection matrix with $n$ dimensional identity matrix $\mathbf{I}_n$. In the following, we are going to use denotation $\mathbf{M}$, $\mathbf{h}$ and $\mathbf{J}_c$ for convenience by omitting dependent variables $\mathbf{q}$ and $\dot{\mathbf{q}}$. 

\subsection{Whole-body control}

In general, a one-step look-ahead problem can be formulated as an optimization problem subject to all physical bounds and environment constraints. In that way, we will lose property analysis of the system since optimization solvers directly give the admissible solutions. By using projection matrix $\mathbf{P}=\mathbf{I}-\mathbf{J}_c^{+}\mathbf{J}_c$ \cite{Aghili2017}\cite{mistry2012operational}, we could decouple Eq. (\ref{e:dynamics}) into constraint-free (Eq. (\ref{e:PSpace})) and constrained (Eq. (\ref{e:orthoSpace})) spaces where we can implement trajectory tracking and satisfy physical constraints respectively. 
\begin{equation}\label{e:PSpace}
\mathbf{P}\mathbf{M}\ddot{\mathbf{q}}+\mathbf{Ph}=\mathbf{PB}\boldsymbol{\tau}
\end{equation}
\begin{equation}\label{e:orthoSpace}
(\mathbf{I}-\mathbf{P})(\mathbf{M}\ddot{\mathbf{q}}+\mathbf{h})=(\mathbf{I}-\mathbf{P})\mathbf{B}\boldsymbol{\tau}+\mathbf{J}_c^{\top}\boldsymbol{\lambda}_c
\end{equation}
Note that Eq. (\ref{e:PSpace}) plus Eq. (\ref{e:orthoSpace}) is equal to the whole system dynamics, i.e., Eq. (\ref{e:dynamics}). However, constraint forces are removed by projector $\mathbf{P}$ in constraint-free space, i.e., Eq. (\ref{e:PSpace}). As pointed out in \cite{aghili2005unified}, the two sub-spaces are not totally decoupled as $\ddot{\mathbf{q}}$ exists in Eq. (\ref{e:orthoSpace}). $\ddot{\mathbf{q}}$ is determined by forward dynamics of Eq. (\ref{e:PSpace}),
\begin{equation}\label{e:forwardDynamics}
    \ddot{\mathbf{q}}=\mathbf{M}_c^{-1}(\boldsymbol{\tau}_m-\mathbf{Ph}+\dot{\mathbf{P}}\dot{\mathbf{q}})
\end{equation}
where $\mathbf{M}_c$ is an invertible matrix defined as $\mathbf{M}_c=\mathbf{PM}+\mathbf{I}-\mathbf{P}$, meanwhile $\boldsymbol{\tau}_m$ is the torque commands generated by a certain control law. Substituting Eq. (\ref{e:forwardDynamics}) into Eq. (\ref{e:orthoSpace}) yields
\begin{equation}\label{e:constraint}
    (\mathbf{I}-\mathbf{P})[\ \mathbf{M}\mathbf{M}_c^{-1}(\boldsymbol{\tau}_m-\mathbf{Ph}+\dot{\mathbf{P}}\dot{\mathbf{q}})+\mathbf{h}]\ =(\mathbf{I}-\mathbf{P})\mathbf{B}\boldsymbol{\tau}+\mathbf{J}_c^{\top}\boldsymbol{\lambda}_c
\end{equation}
Then we formulate the following Quadratic Programming problem in constrained space,
\begin{equation}\label{e:qp}
    \begin{aligned}
        & \underset{\boldsymbol{\tau}_c, \boldsymbol{\lambda}_c}{\text{minimize}}
        & & \frac{1}{2}\|\boldsymbol{\tau}_c\|_2^2 \\
        & \text{subject to}
        & & \text{Eq.} (\ref{e:constraint}) \\
        &&&  \boldsymbol{\tau}_{\text{min}}-\boldsymbol{\tau}_m\leq\boldsymbol{\tau}_c\leq\boldsymbol{\tau}_{\text{max}}-\boldsymbol{\tau}_m\\
        &&& \begin{bmatrix}0 & 0 -1\end{bmatrix}\boldsymbol{\lambda}_{c,i} &\leq 0 \\
        &&& \begin{bmatrix} 1 & 0 & \small{-\frac{1}{2}}\sqrt{2}\mu\end{bmatrix}\text{abs}(\boldsymbol{\lambda}_{c,i}) &\leq 0 \\
        &&& \begin{bmatrix} 0 & 1 & \small{-\frac{1}{2}}\sqrt{2}\mu\end{bmatrix}\text{abs}(\boldsymbol{\lambda}_{c,i}) &\leq 0
    \end{aligned} 
\end{equation}
where $\boldsymbol{\tau}_c$ denotes $(\mathbf{I}-\mathbf{P})\mathbf{B}\boldsymbol{\tau}$ of Eq. (\ref{e:orthoSpace}), and $\boldsymbol{\lambda}_{c,i} \in \mathbb{R}^3$ denotes the contact force of $i$th contact point, $\mu$ is the friction coefficient. The last three inequality constraints in Eq. (\ref{e:qp}) represent the linearized friction cone constraints with 4-edge pyramid.
The final torque command generated from our controller is the sum of both sub-spaces
\begin{equation}
    \boldsymbol{\tau}=\boldsymbol{\tau}_m+\boldsymbol{\tau}_c^*
\end{equation}
To generate $\boldsymbol{\tau}_m$, people can resort to normal robotic arm controllers with the consideration of under-actuation factor $\mathbf{B}$ in Eq. (\ref{e:PSpace}). We derived the Cartesian impedance control law in \cite{Xin2018} as follows,
\begin{equation}\label{e:tau_m}
    \boldsymbol{\tau}_m=(\mathbf{PB})^{+}\mathbf{P}(\mathbf{J}_s^{\top}\mathbf{F}_s+\mathbf{N}_s\mathbf{J}_b^{\top}\mathbf{F}_b)
\end{equation}
along with
\begin{equation}\label{e:control_law}
    \mathbf{F}_i=\mathbf{h}_{c,i}+\boldsymbol{\Lambda}_{c,i}\mathbf{\ddot{x}}_{d,i}-\mathbf{D}_{d,i}\dot{\tilde{\mathbf{x}}}_{i}-\mathbf{K}_{d,i}\tilde{\mathbf{x}}_i, \quad i=s \text{ or }b
\end{equation}
where $\mathbf{F}_i$ denotes the operational space control command which enforces the system to obey the impedance behavior subject to external disturbances,
\begin{equation}\label{e:impedance}
    \boldsymbol{\Lambda}_{c,i}\ddot{\tilde{\mathbf{x}}}_{i}+\mathbf{D}_{d,i}\dot{\tilde{\mathbf{x}}}_{i}+\mathbf{K}_{d,i}\tilde{\mathbf{x}}_i=\mathbf{F}_{x,i}, \quad i=s \text{ or }b
\end{equation}
where $\tilde{\mathbf{x}}_i=\mathbf{x}_i-\mathbf{x}_{i,d}$ is the deviation of end-effectors in Cartesian space. Subscripts $s$ and $b$ denote swing feet and the base respectively. $\mathbf{h}_{c,i}$ and $\boldsymbol{\Lambda}_{c,i}$ represent nonlinear term of operational space dynamics and operational space inertia matrix (see \cite{Xin2018} for details). $\mathbf{N}_s$ is the dynamic consistent null-space projector \cite{mistry2012operational} of the swing foot, which enforces strictly hierarchical priorities. In the case of foot posture control, $\mathbf{N}_s$ will deal with the overlap between swing foot Jacobian $\mathbf{J}_s$ and base Jacobian $\mathbf{J}_b$, leading to the convenience of leaving base Jacobian $\mathbf{J}_b$ to be always a $\mathbb{R}^{6\times(n+6)}$ matrix. As the torso is in the null-space of the swing foot, the torso is enforced to satisfy the swing foot's motion requirement, which results in automatic motion coordination and reachability extension of foot. For locomotion control, we only control three dimensions of a foot, leading to $\mathbf{J}_s \in \mathbb{R}^{3\times(n+6)}$. On the other hand, for manipulation tasks, we control six dimensions of a foot (in fact, the shank) with $\mathbf{J}_s \in \mathbb{R}^{6\times(n+6)}$. 

One of the benefits of our loco-manipulation controller is that we can use Eq. (\ref{e:impedance}) to estimate the external forces, such as contact forces acting on the foot during doing manipulation tasks. Note that the torso motion error does not affect the foot position error because the
torso is controlled in the null-space of the manipulation foot. We can always only
measure the motion of the foot and then use the following equation to estimate
external force $\mathbf{F}_x$ without any torso motion error interfering:
\begin{equation}\label{e:estimation}
    \hat{\mathbf{F}}_x=\boldsymbol{\Lambda}_{c,s}\ddot{\tilde{\mathbf{x}}}_{s}+\mathbf{D}_{d,s}\dot{\tilde{\mathbf{x}}}_{s}+\mathbf{K}_{d,s}\tilde{\mathbf{x}}_s.
\end{equation}
Here in our experiments, we employ this
estimation as haptic feedback for teleoperation, and thus do not require a force/torque sensor at the point of contact.

\section{EVALUATION}
\subsection{Whole-body motion with fixed end-effector position and orientation}
\begin{figure}[t!]
    \centering
    \includegraphics[width=.95\linewidth]{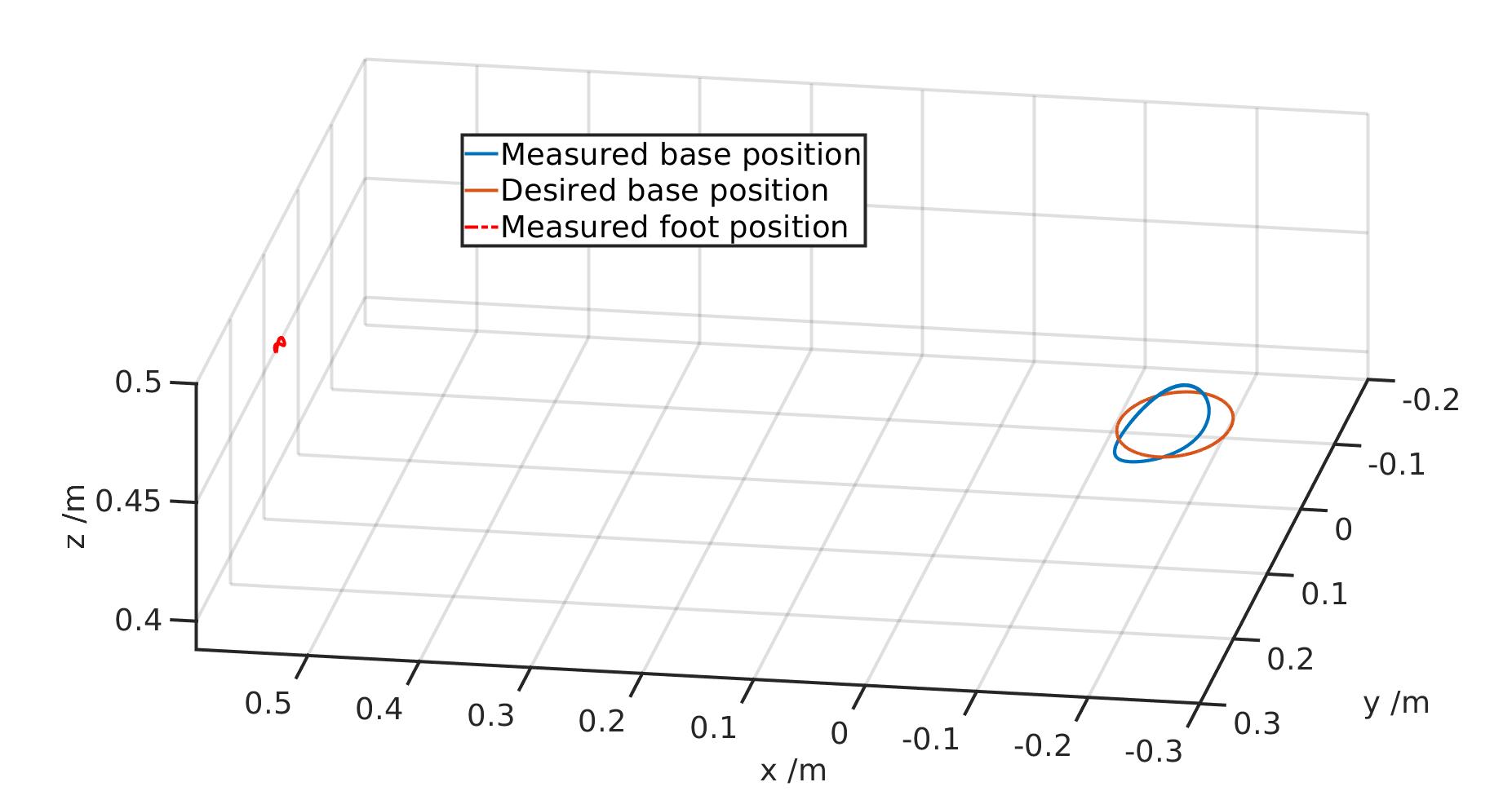}
    \caption{Positions of the base and the non-contact foot while moving the base to track a circular trajectory with fixed end-effector posture.}
    \label{fig:positions}
\end{figure}

\begin{figure}[t!]
    \centering
    \includegraphics[width=.95\linewidth]{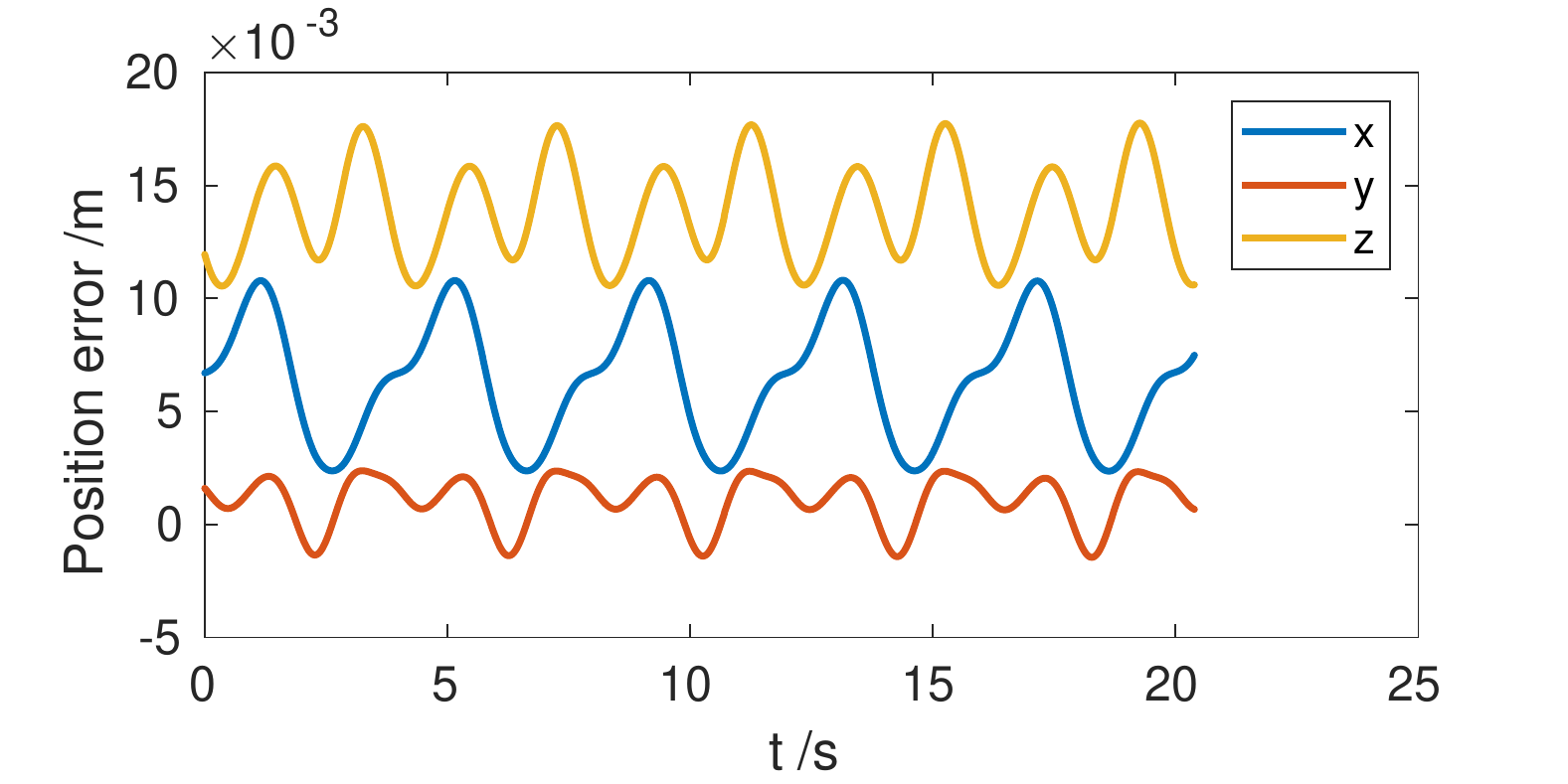}
    \caption{Position error of the non-contact foot during whole-body control verification.}
    \label{fig:positionError}
\end{figure}

\begin{figure}[t!]
    \centering
    \includegraphics[width=.95\linewidth]{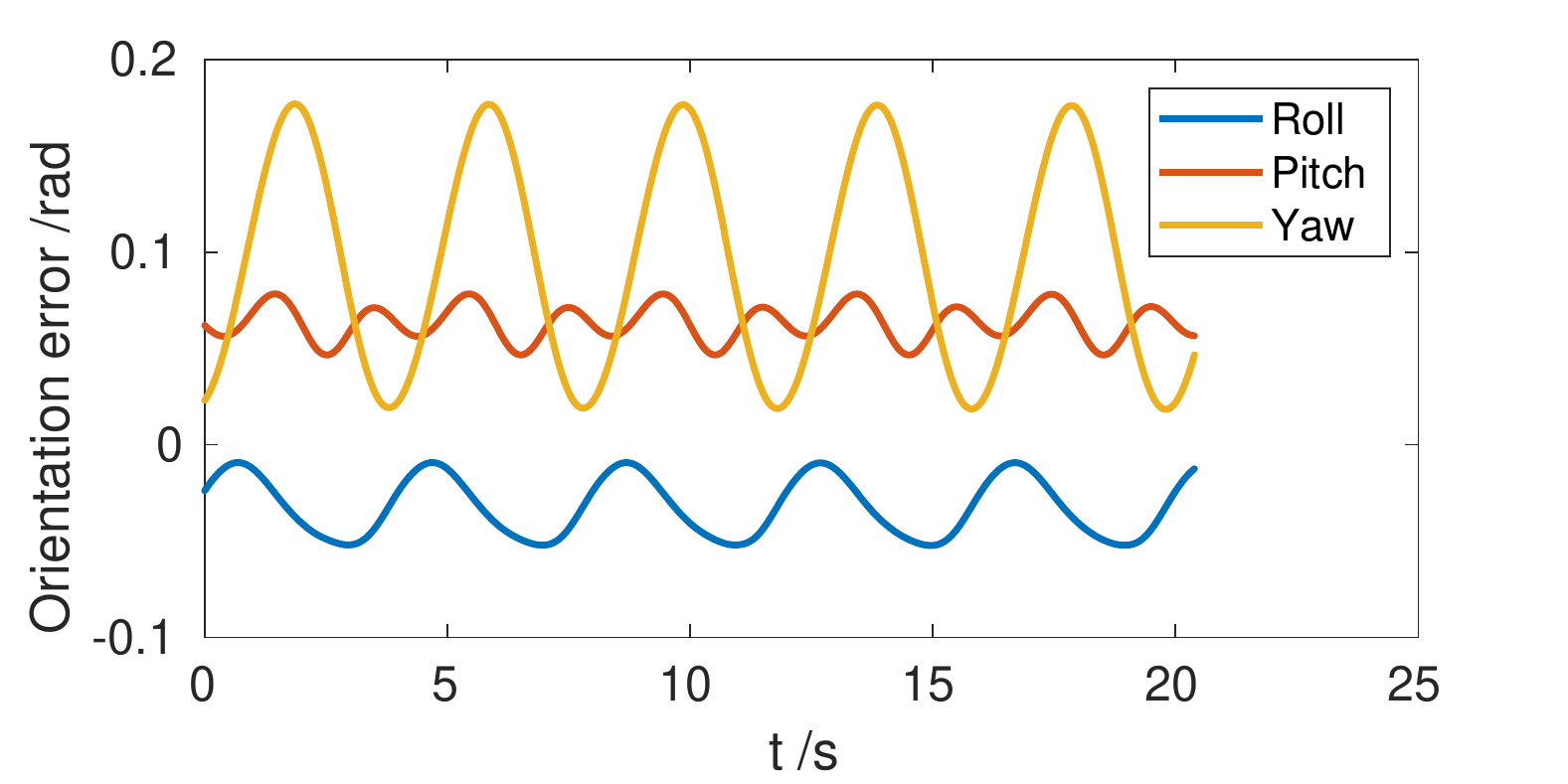}
    \caption{Orientation error of the non-contact foot during whole-body control verification.}
    \label{fig:orientationError}
\end{figure}

\begin{figure}[t!]
\centering
\begin{overpic} [width=.5\linewidth]{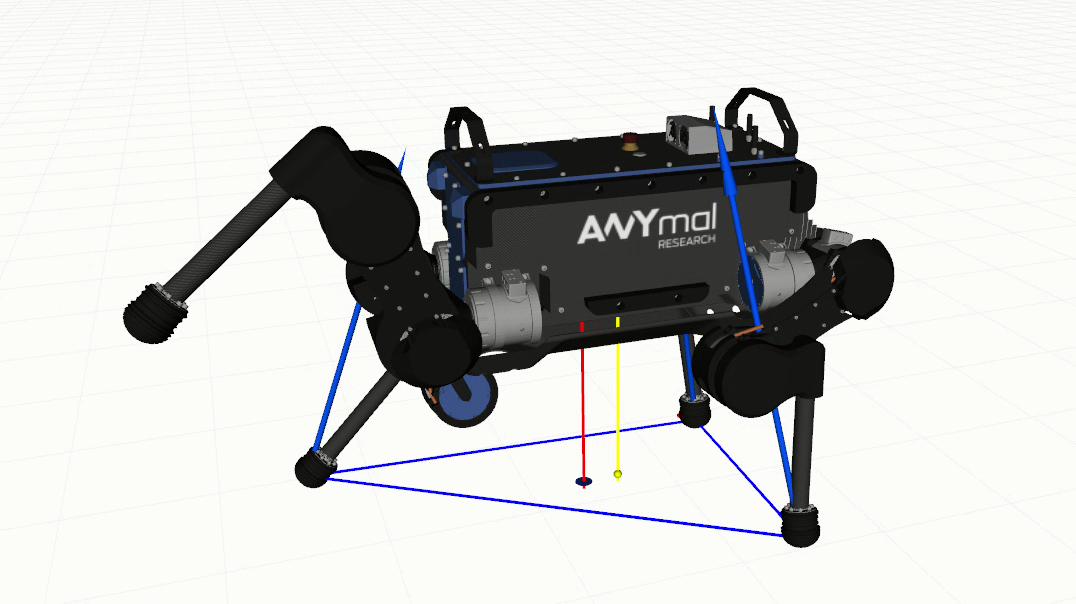}\put(5,5){\manuallabel{f:chicken_roll1}{(a)}\ref{f:chicken_roll1}}\end{overpic}~
\begin{overpic} [width=.5\linewidth]{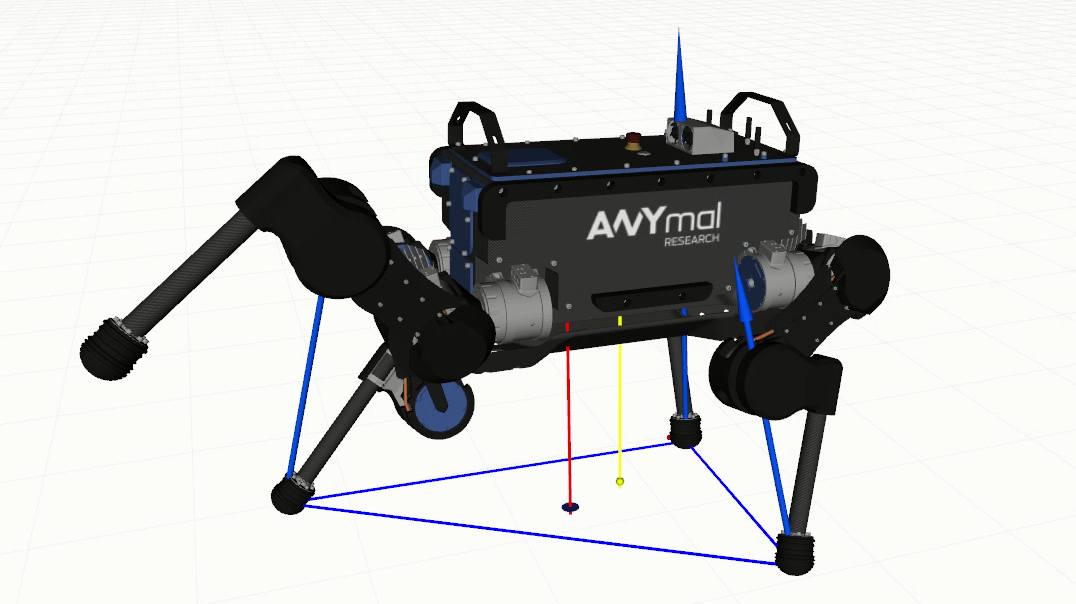}\put(5,5){\manuallabel{f:chicken_roll2}{(b)}\ref{f:chicken_roll2}} \end{overpic}\\
\begin{overpic} [width=.5\linewidth]{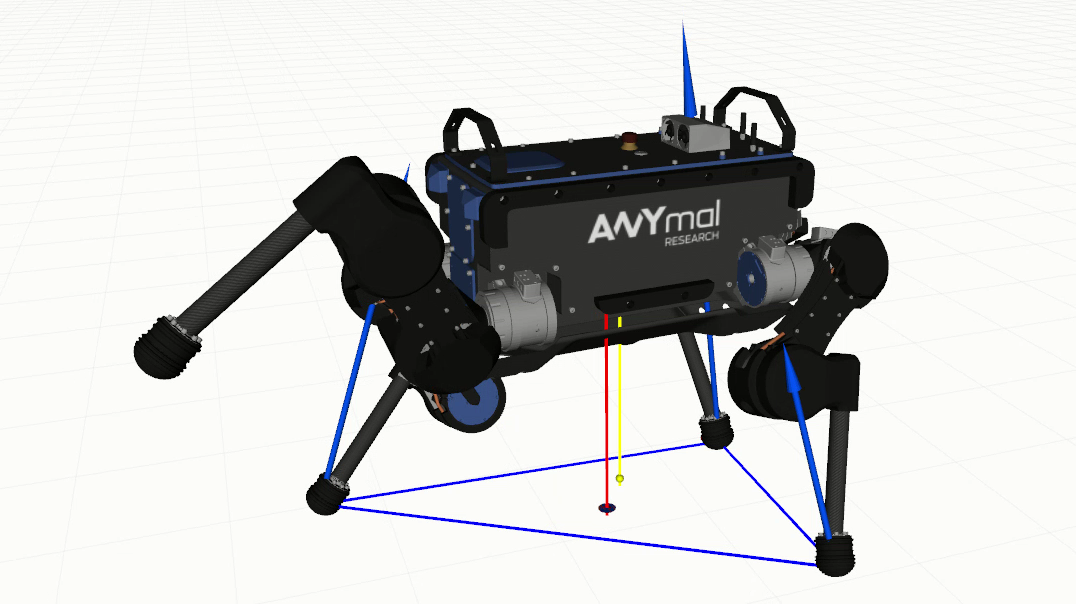}\put(5,5){\manuallabel{f:chicken_roll3}{(c)}\ref{f:chicken_roll3}}\end{overpic}~
\begin{overpic} [width=.5\linewidth]{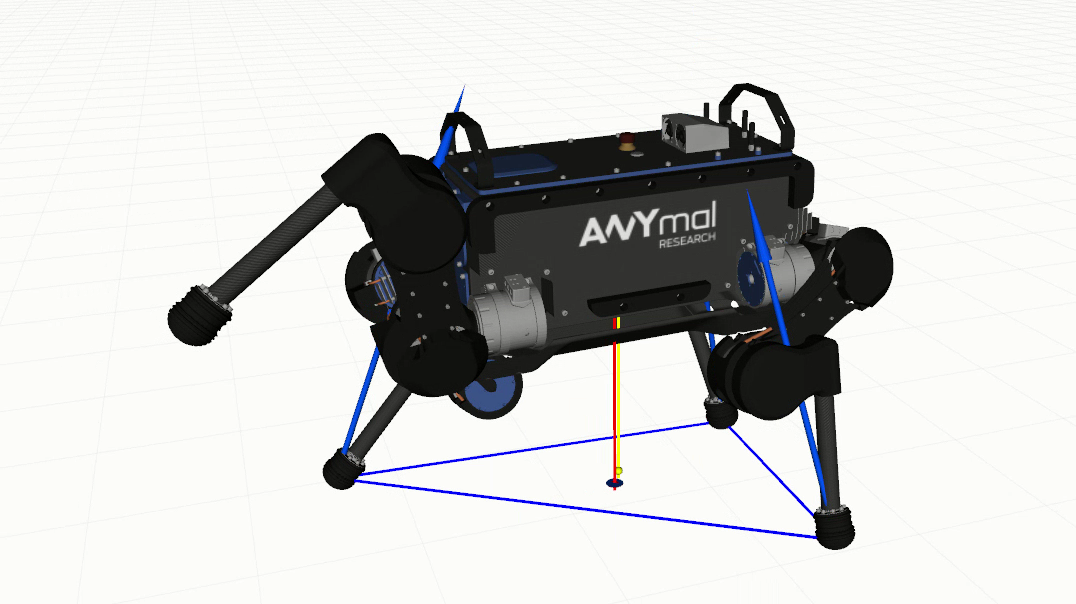}\put(5,5){\manuallabel{f:chicken_roll4}{(d)}\ref{f:chicken_roll4}} \end{overpic}
\\\vspace{-1mm} 
\caption{Whole-body control verification: base tracks a circular trajectory while maintaining left-fore foot's current posture.}  
\label{f:chicken_roll} 
\end{figure}

The primary feature of a whole-body controller is to handle multiple tasks. A typical test for whole-body control is to maintain one end-effector (a foot) with fixed posture while moving another end-effector (the base), as shown in Fig. \ref{f:chicken_roll}. In this simulation, we controlled 5 dimensions of the left-fore foot with respect to the fixed inertia frame by relaxing Yaw of the foot. Fig. \ref{fig:positions} shows the positions of the torso and the left-fore foot during this simulation. The torso was tracking a circle in x-y plane with desired constant height. Figs. \ref{fig:positionError} and \ref{fig:orientationError} depict the foot's position error and orientation error during the simulation respectively. It is obvious that the error of Yaw is much greater than the other directions whereas the tracking performance of torso is not good as the foot. That is reasonable since the torso control is in the null-space of foot control. It proves that the null-space projector $\mathbf{N}_s$ in Eq. (\ref{e:tau_m}) works as expected. Readers can check out this demonstration motion in the accompanying video (\url{https://youtu.be/tIyfUjJgJIM}). 

\subsection{Body-ground contact evaluation}
\begin{figure}[t!]
\centering
\begin{overpic} [width=.5\linewidth]{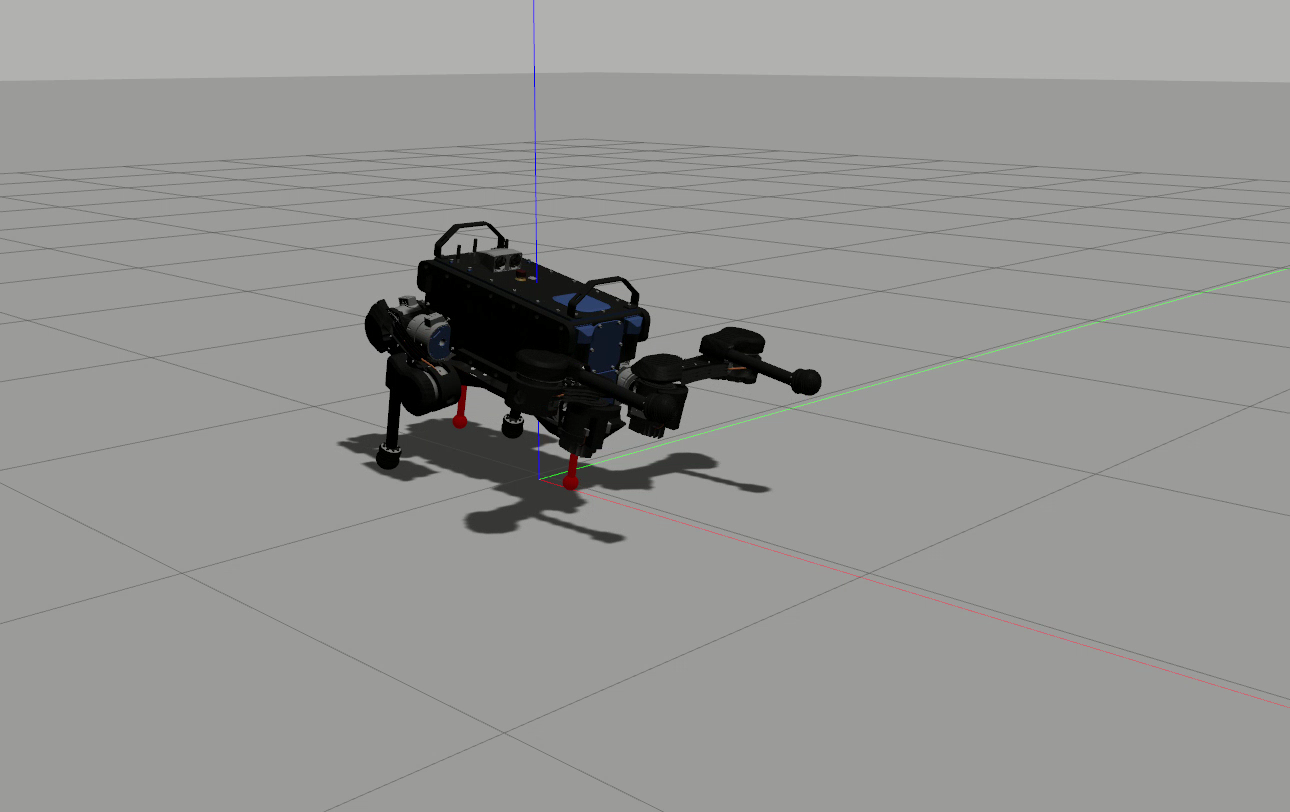}\put(5,5){\manuallabel{f:roll_torso1}{(a)}\ref{f:roll_torso1}}\end{overpic}~
\begin{overpic} [width=.5\linewidth]{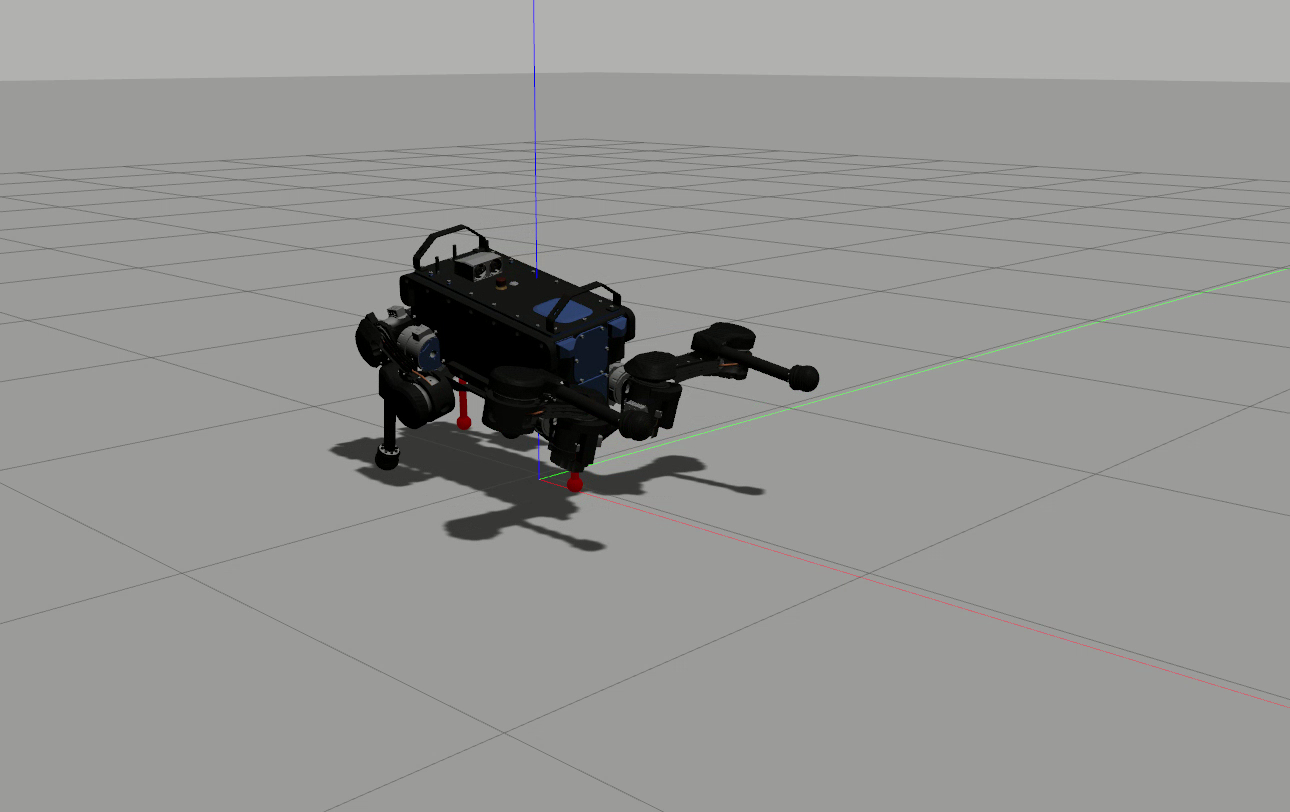}\put(5,5){\manuallabel{f:roll_torso2}{(b)}\ref{f:roll_torso2}} \end{overpic}\\ \vspace{1mm}
\begin{overpic} [width=.5\linewidth]{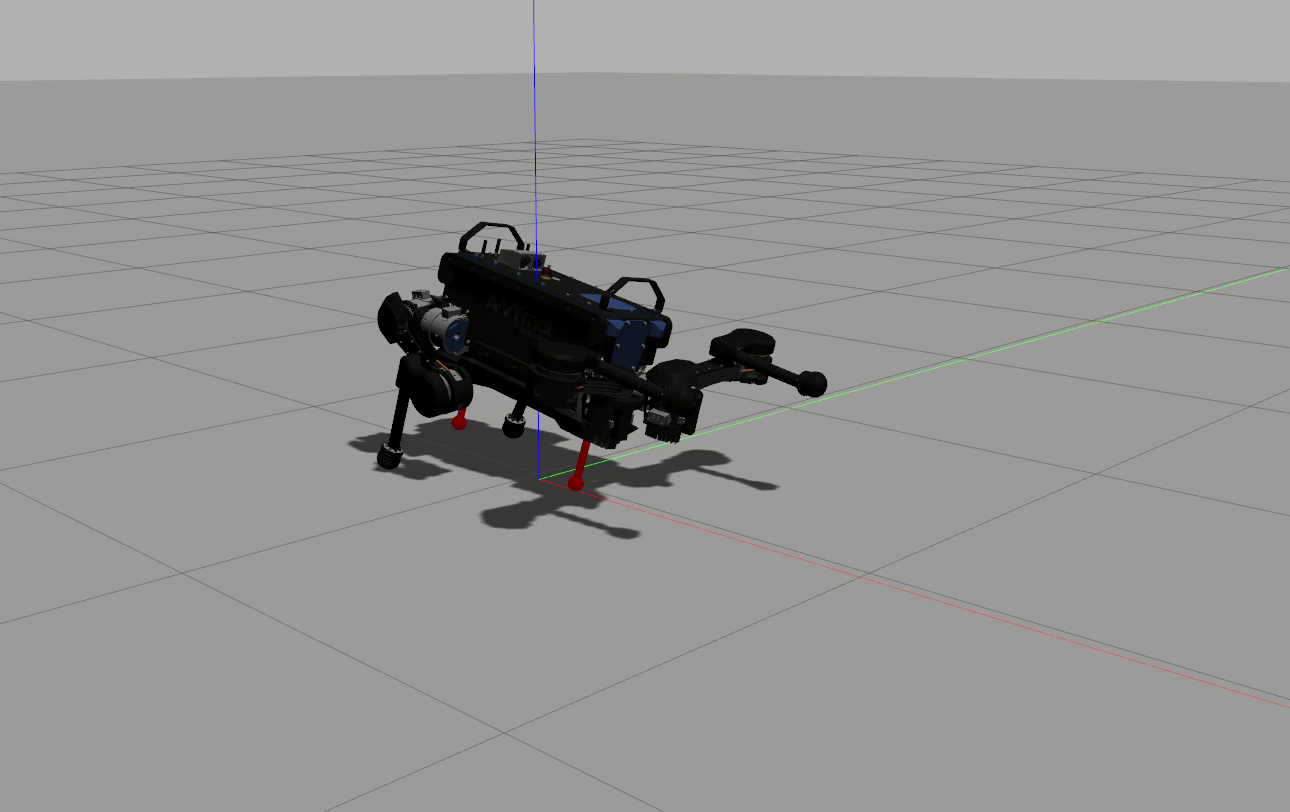}\put(5,5){\manuallabel{f:roll_torso3}{(c)}\ref{f:roll_torso3}}\end{overpic}~
\begin{overpic} [width=.5\linewidth]{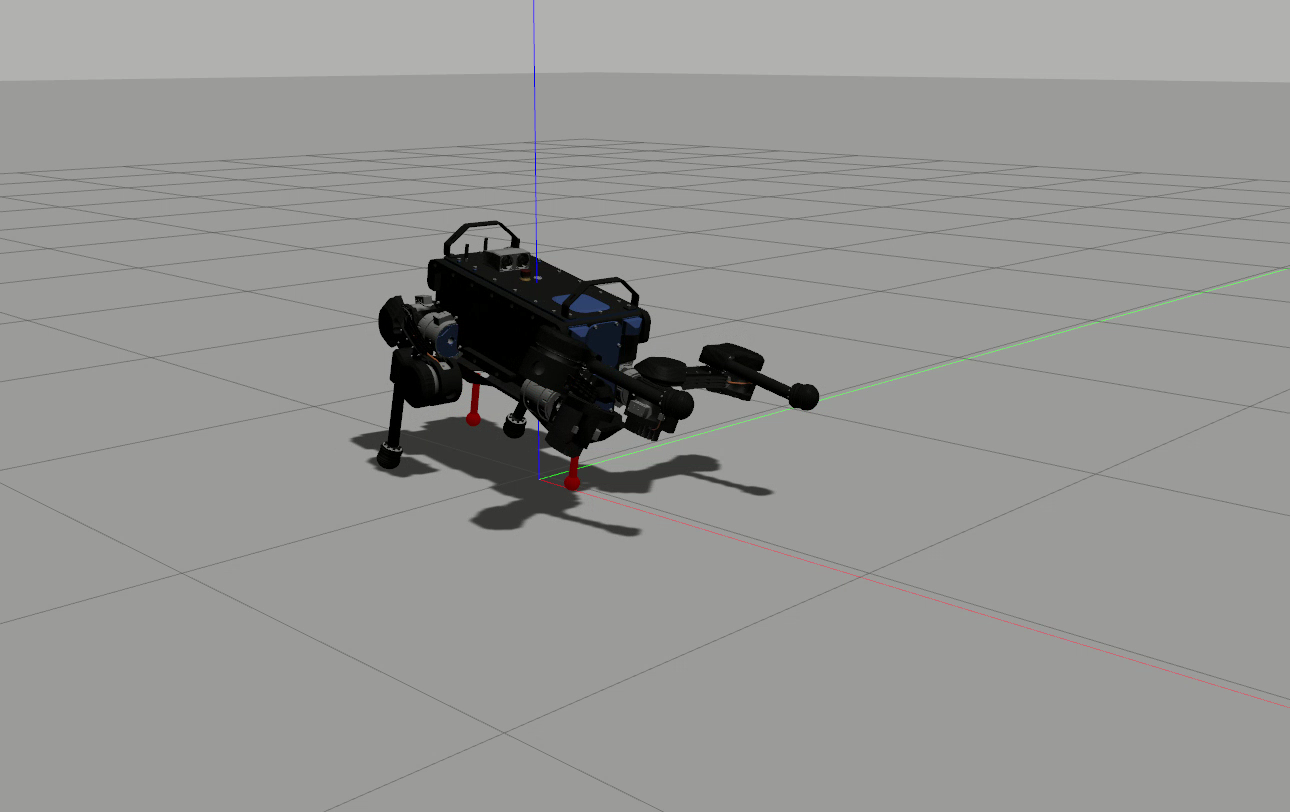}\put(5,5){\manuallabel{f:roll_torso4}{(d)}\ref{f:roll_torso4}} \end{overpic}
\\\vspace{-1mm} 
\caption{Moving orientation of the torso supporting by one prong and two rear legs: (a) rotating Roll; (b) rotating Yaw.}  
\label{f:rollTorso} 
\end{figure}

In this subsection, we evaluate the controller can handle body-ground contact cases. In Fig. \ref{f:rollTorso}, one prong and two rear feet get contact with ground. The constraint Jacobian $\mathbf{J}_c=\begin{bmatrix}\mathbf{J}_p^\top & \mathbf{J}_{rh}^\top &\mathbf{J}_{lh}^\top\end{bmatrix}^\top \in \mathbb{R}^{9 \times 18}$ where subscripts $p$, $rh$ and $lh$ stand for prong, right-hind and left-hind respectively. $\mathbf{J}_p$ is a constant matrix since it is not configuration dependant. In this case, the torso can rotate in three dimensions around the prong actuated by two rear legs. The proposed controller does not need to adapt to this special case because the projection matrix $\mathbf{P}$ exists as usual. We defined a sequence of rotation actions to move Roll, Pitch and Yaw and implemented in simulation as shown in Fig. \ref{f:rollTorso}. It is also included in the accompanying video. During the orientation movement, we control the two fore feet to keep current positions, which is achieved by the hierarchical feature of our controller as discussed in former subsection. Furthermore, when there are two prongs contacting with ground, $\mathbf{J}_c$ is not full row rank because two prongs are linear dependant. $Rank(\mathbf{J}_c \in \mathbb{R}^{12 \times 18}) = 11$ in this case. We use singular value decomposition (SVD) to compute pseudo-inverse of $\mathbf{J}_c$ with zero or near-zero singular values removed, leading to the projector $\mathbf{P}$.

\subsection{Testing in industrial environments}

\begin{figure}[t!]
\centering
\begin{overpic} [width=.5\linewidth]{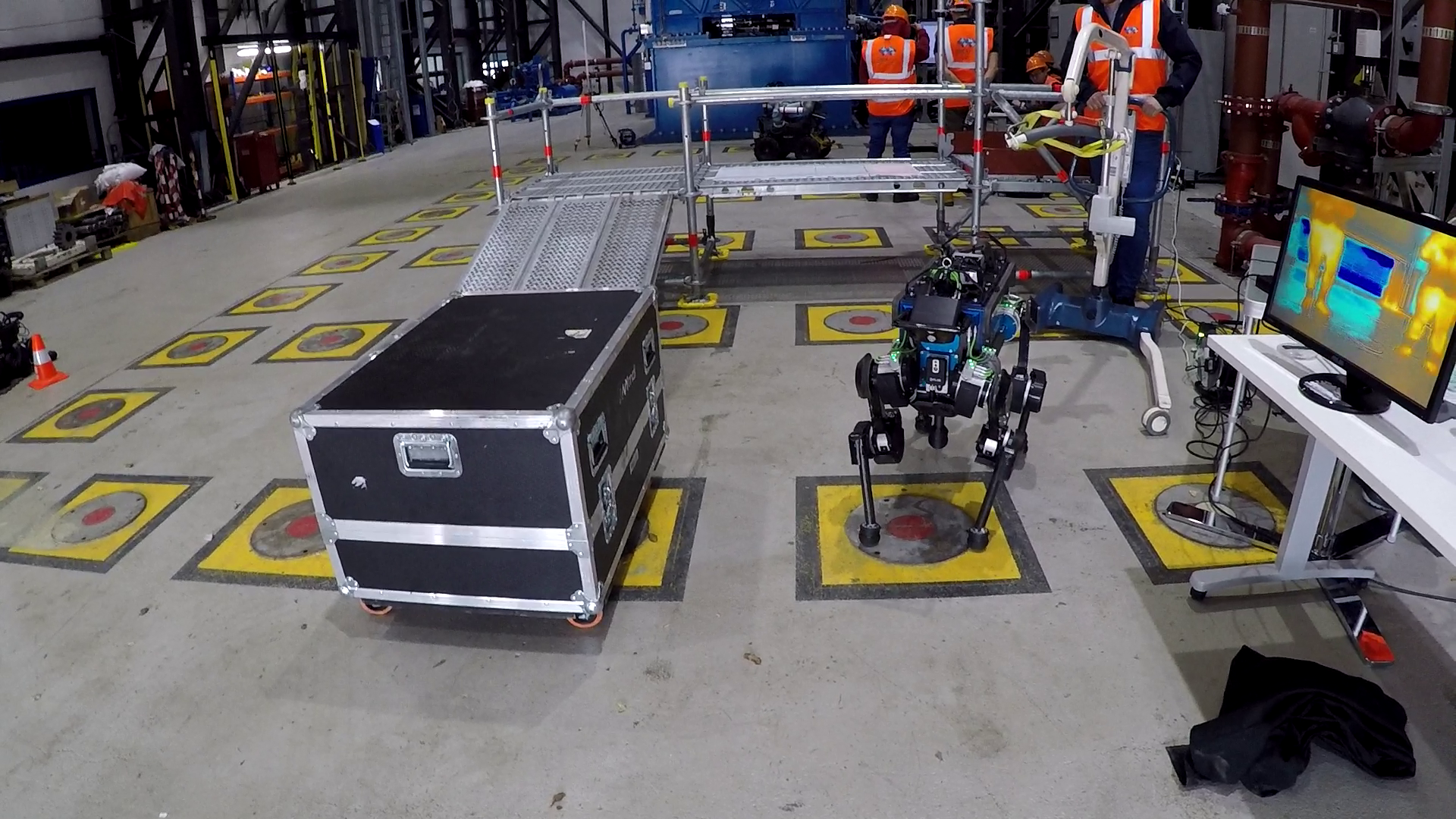}\put(5,5){\manuallabel{f:orca_test1}{(a)}\ref{f:orca_test1}} \end{overpic}~
\begin{overpic} [width=.5\linewidth]{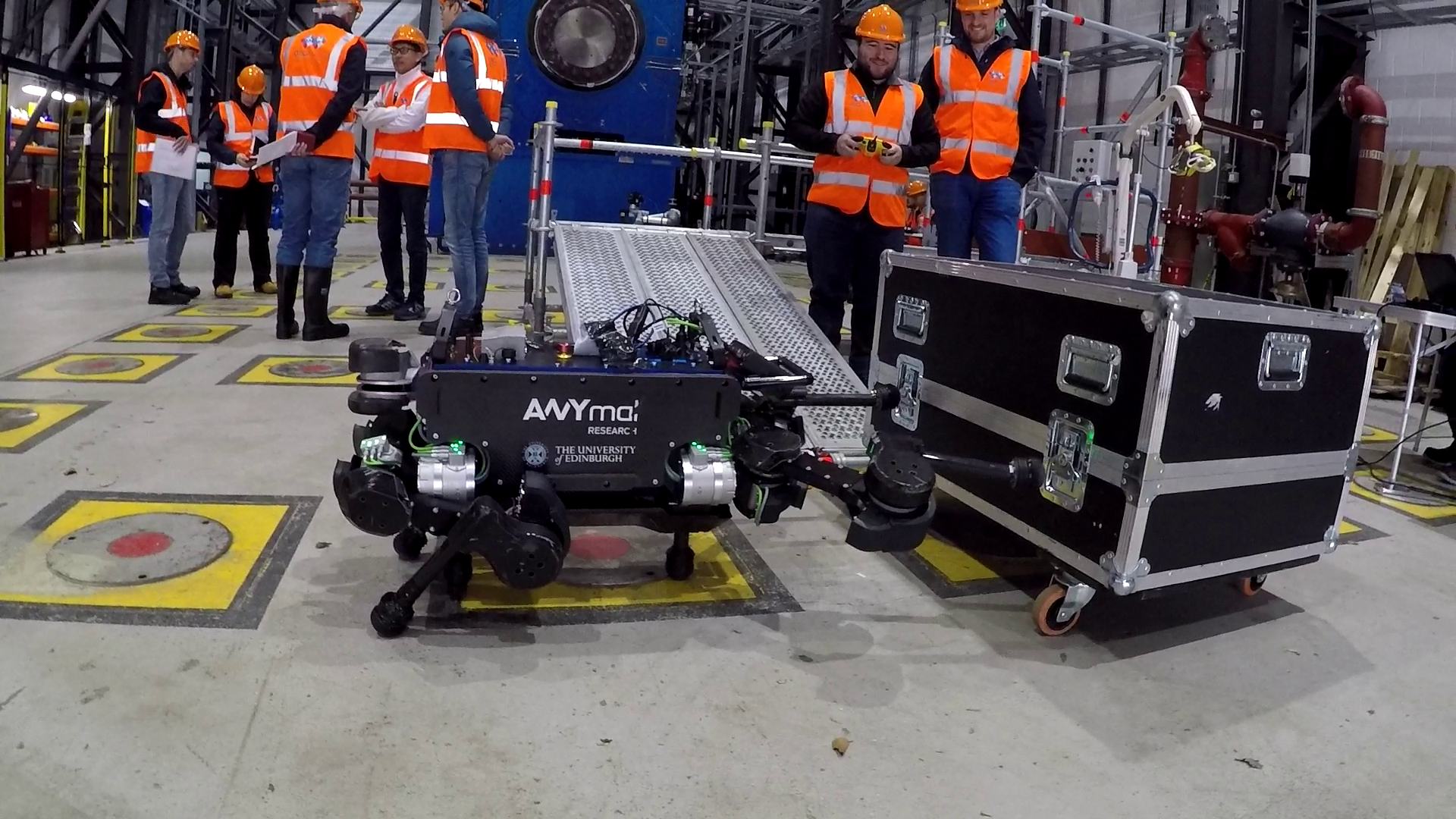}\put(5,5){\manuallabel{f:orca_test2}{(b)}\ref{f:orca_test2}} \end{overpic}\\  \vspace{1mm}
\begin{overpic} [width=.5\linewidth]{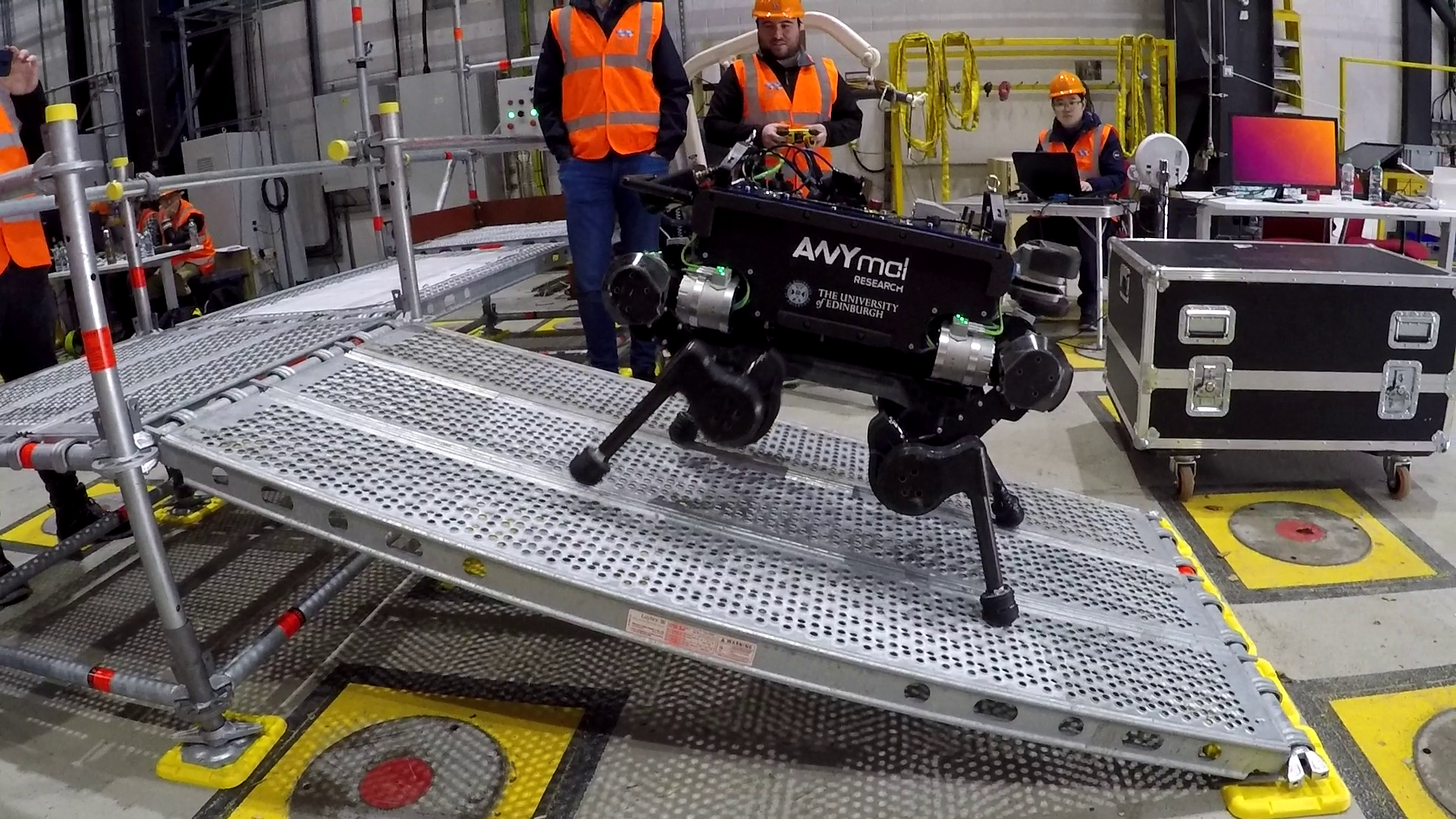}\put(2,2){\manuallabel{f:orca_test3}{(c)}\ref{f:orca_test3}}\end{overpic}~
\begin{overpic} [width=.5\linewidth]{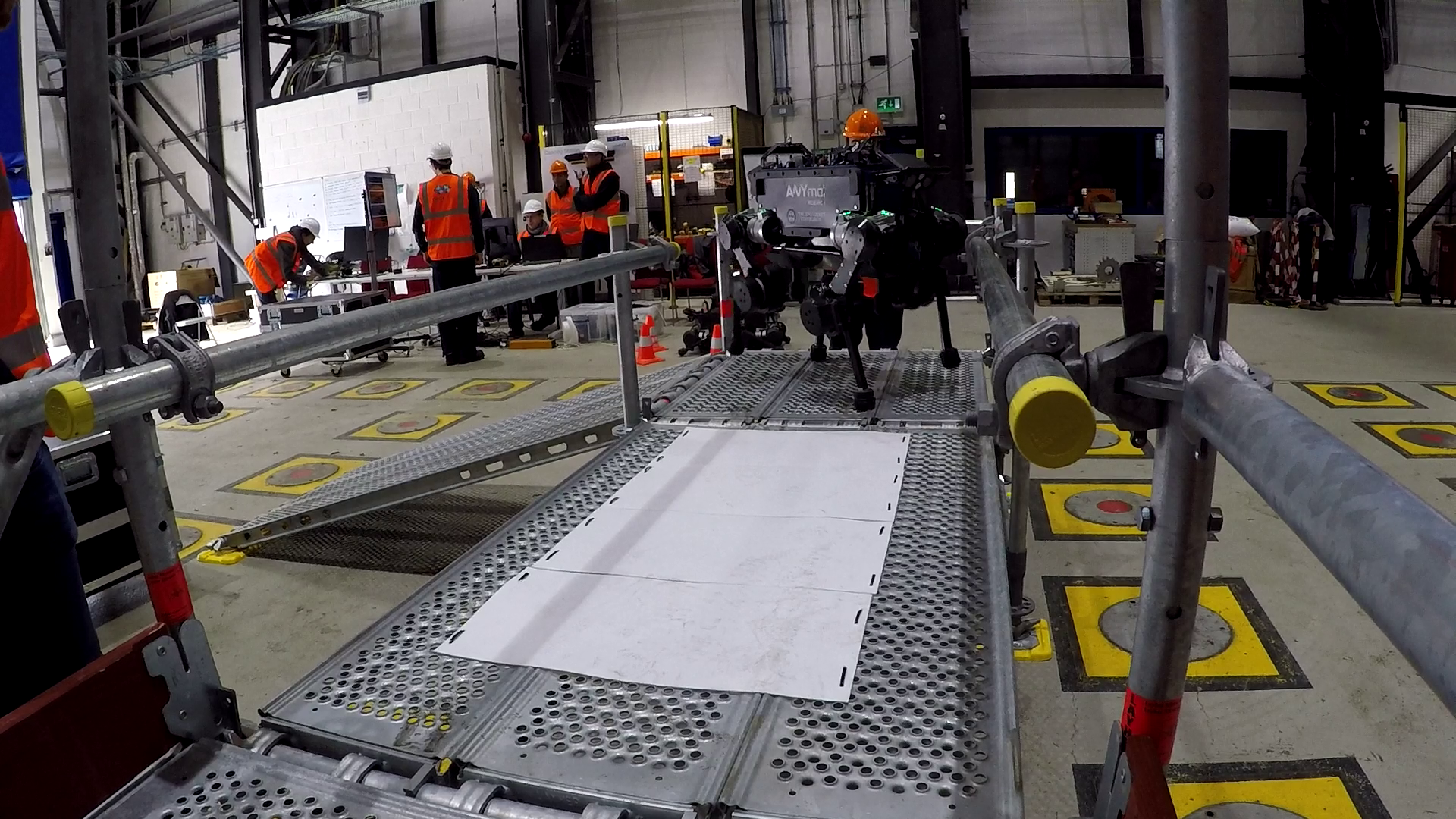}\put(15,2){\manuallabel{f:orca_test4}{(d)}\ref{f:orca_test4}} \end{overpic}\\ \vspace{1mm}
\begin{overpic} [width=.5\linewidth]{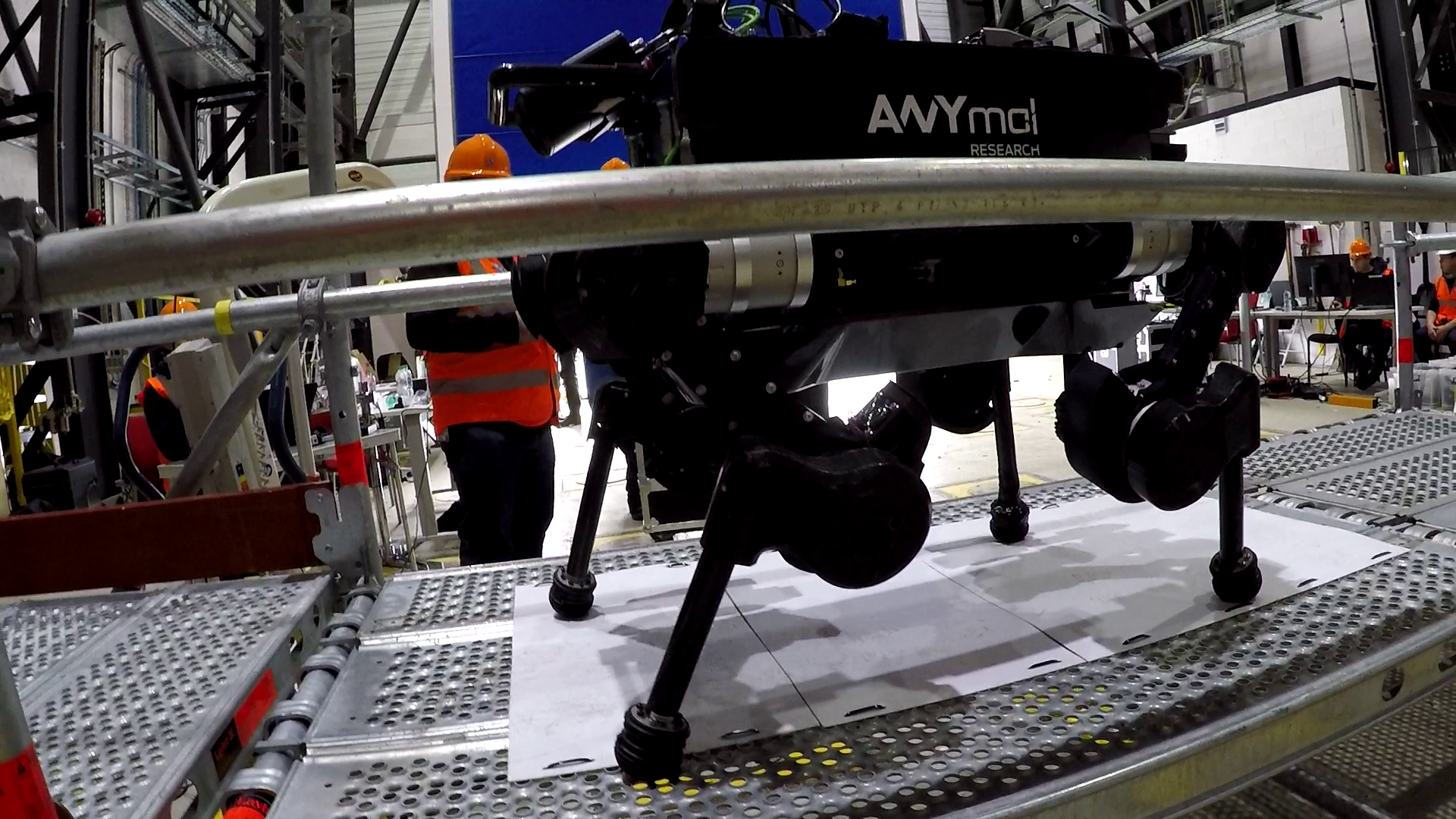}\put(5,5){\manuallabel{f:orca_test5}{(e)}\ref{f:orca_test5}}\end{overpic}~
\begin{overpic} [width=.5\linewidth]{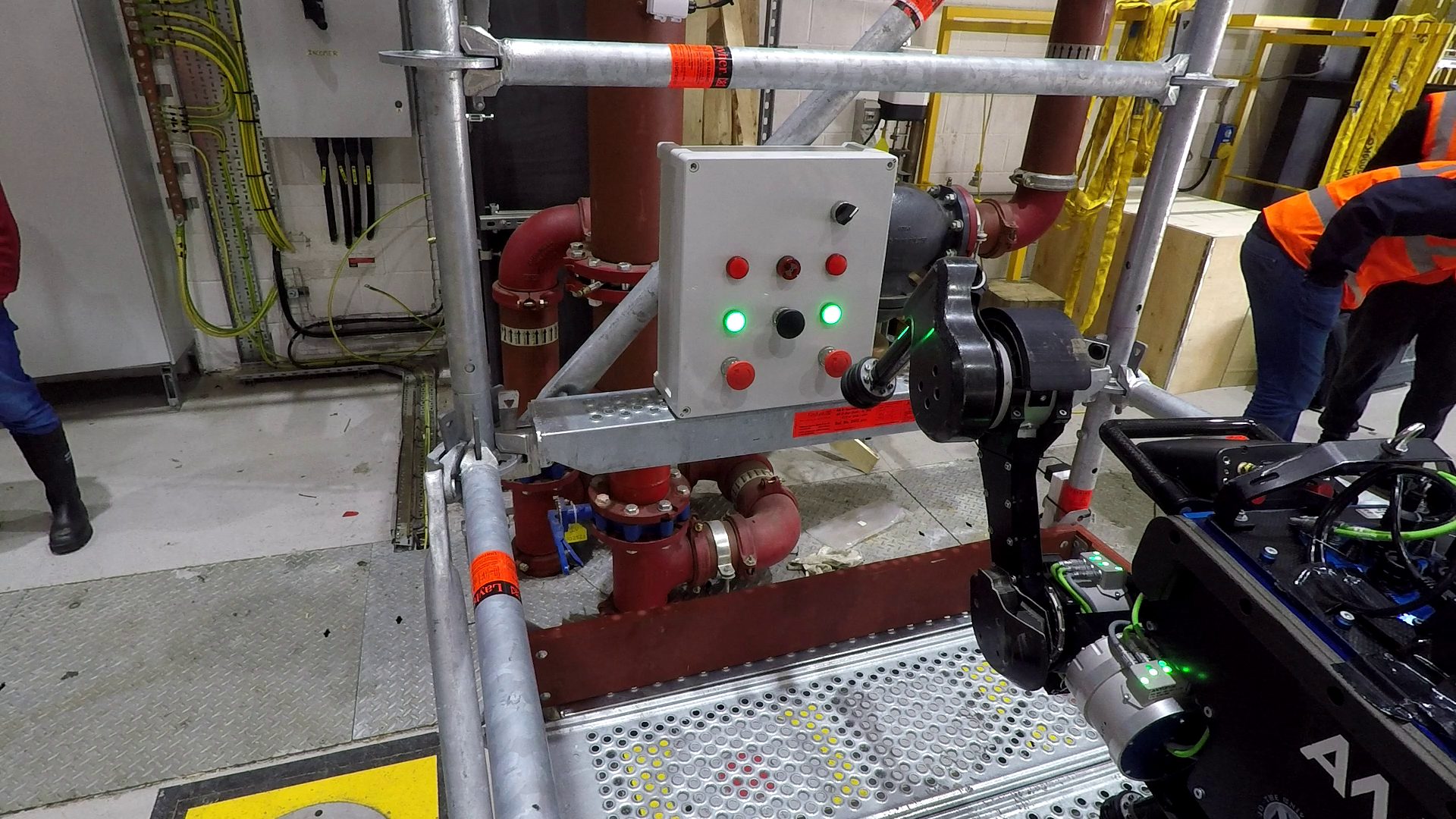}\put(5,5){\manuallabel{f:orca_test6}{(f)}\ref{f:orca_test6}} \end{overpic}
\\\vspace{-1mm} 
\caption{Testing our system and algorithms in an industrial site: (a) starting from origin; (b) pushing the obstacle away; (c) climbing up on a \SI{20}{\degree} slope; (d) turning around; (e) walking over slippery terrain; (f) teleoperating a foot to press E-stop.}  
\label{f:testing} 
\end{figure}

The ORCA project gave us an industrial environment to test our work as described in the Introduction. Our industrial partners (such as Total, BP and their sub-contractors) designed a mock industrial application scenario for us. The robot equipped with cameras is deployed to press down an E-stop button on a platform. The robot and operator were sharing autonomy via command and feedback information communication. The operator used an HRI remote controller to steer the robot for locomotion. As shown in Fig. \ref{f:testing}, the robot had to climb up a slope and walk over a slippery bridge covered by artificial ice before approaching the electrical box where E-stop button located. However, we intentionally placed a large box in front of the slope. The robot trotted to a side of the box and lay on the two prongs and two rear legs in order to use two front legs to push the box away. Then the robot switched to a static-walking gait to climb the \SI{20}{\degree} slope and also to traverse the slippery terrain. A state estimator \cite{bloesch2013state} updates the inclined terrain information for the control loop. After climbing up on the platform, the operator switches to a trotting gait to speed up turning \SI{90}{\degree} in place. While traversing the bridge, the operator had to change the friction coefficient parameter in the controller manually before stepping on the artificial ice since there was no algorithms to estimate the friction coefficient online. The friction coefficient between robot's feet and the artificial ice is about 0.2. That is close to some terrains covered by water and oil in industrial environments. Eventually, the operator teleoperated the left-fore foot to reach the E-stop button and pressed it down with the assistant of camera vision. During lifting the foot, the leg overcame a singularity configuration without instability, which proves the robustness of our controller. The total process was recorded in the accompanying video.

The success of this application experiment gives us a baseline of our system's capability. The task can be easily modified for similar scenarios by changing any of the sub-problems to fit the needs. The advantage of using the whole-body controller in this scenario is that the framework can handle all the locomotion and operation modes, which allows us to execute different tasks by switching motion trajectories and foot contact sequences.

\section{Discussion}

We have presented an architecture for whole-body control and planning of legged robots. This system exploits a generic formulation of tasks as a QP-based semi-analytical operational space controller, and it integrates inputs from normal joystick, haptic joystick, perception, and motion planner. The Cartesian impedance controller then minimizes the tracking error and also maintains compliance against disturbances while satisfying physical constraints. The formulation of the problem allows us to adapt to a variety of loco-manipulation tasks and constraints including body-ground singular constraints. Particularly, the body-ground contact case is helpful for saving energy, improving payload capability as well as releasing more DOFs for manipulation.  

Our evaluation on the whole-body maneuver validates the architecture. The tracking results then show the overall performance of the system. The rotating of base around body-ground contact point validates the generality of the whole-body controller. The last experiment host in the industrial environment strengthen the practicality of employing quadruped robots for real world needs. Equipped with various sensors, the robot can be useful for inspection and intervention and reduce workers' labour in extreme environments. Such an approach can rapidly accelerate the development and deployment of robotic systems in autonomous nuclear equipment maintenance, off-shore asset maintenance, and many other fields.

However, the system also has some aspects that can be improved in the future. The foot cannot reach very high. The E-stop button in our experiment is \SI{0.8}{\meter} high. Therefore, the application of teleoperating a foot is limited to low height operation, such as improving perception accuracy via haptic foot exploration. For more complex manipulation tasks, we need an extra arm mounted on the torso to increase DOFs and reachability \cite{bellicoso2019alma}. Second, the motion planner is a one step looking forward planner. A long time horizon considering planner could improve efficiency and stability of the robot because it will make full use of the robot's physical capability \cite{winkler2018gait}. The controller proposed in this paper is available for some dynamic gaits such as walking trotting and flying trotting. However dynamic gaits will rely heavily on suitable foot step planning to keep balance. 

\bibliographystyle{IEEEtran}
\bibliography{IEEEtranBST/bibliography}
\end{document}